\documentclass{article}

\usepackage{PRIMEarxiv}

\usepackage[utf8]{inputenc} 
\usepackage[T1]{fontenc}    
\usepackage{hyperref}       
\usepackage{url}            
\usepackage{booktabs}       
\usepackage{amsfonts}       
\usepackage{nicefrac}       
\usepackage{microtype}      
\usepackage{lipsum}
\usepackage{fancyhdr}       
\usepackage{graphicx}       
\graphicspath{{media/}}     

\title{Research on color recipe recommendation based on unstructured data using TENN
}

\author{
  Seongsu Jhang, Donghwi Yoo, Jaeyong Kown \\
  ICT Convergence Research Center \\
  KETI \\
  Changwon City, South Gyeongsang Province\\
  \texttt{\{jss0221, elight, jykwon\}@keti.re.kr} \\
}

\begin{document}
\maketitle

\begin{abstract}
Recently, services and business models based on large language models (LLM), such as OpenAI's Chat-GPT, Google's BARD, and Microsoft's copilot, have been introduced, and the applications utilizing natural language processing with deep learning are increasing, and it is one of the natural language pre-processing methods. Conversion to machine language through tokenization and processing of unstructured data are increasing. Although algorithms that can understand and apply human language are becoming increasingly sophisticated, it is difficult to apply them to processes that rely on human emotions and senses in industries that still mainly deal with standardized data. In particular, in processes where brightness, saturation, and color information are essential, such as painting and injection molding, most small and medium-sized companies, excluding large corporations, rely on the tacit knowledge and sensibility of color mixers, and even customer companies often present non-standardized requirements. . In this paper, we proposed TEDNN(Tokenizing Encoder Deep Neural Network to infer color recipe based on unstructured data with emotional natural language, and demonstrated it.

\end{abstract}


\section{Introduction}
Recently, artificial neural network-based algorithms are being rapidly absorbed into the various industries due to the rapid development of hardware like GPU that enables efficient parallel computation. Through this, productivity is increasing at a rapid pace in advanced and manufacturing processes from various perspectives, such as industrial AI and autonomous manufacturing.
However, the basis for these algorithms and advancements must be preceded by digitalization. As for the development of learning models using big data, SMEs that have not yet prepared the conditions are lagging behind global conglomerates.
Regarding the tacit knowledge and know-how in the process, data acquisition is difficult, which is why the majority of manufacturing companies like SMEs remain at an automation level. 
\\ 
\null\quad In particular, many processes that involve color engineering, such as painting, synthetic resin production, paint manufacturing, and injection molding, heavily rely on the intuition and tacit knowledge of color technicians(color engineer), making it challenging to digitize this knowledge. Furthermore, some companies treat this information as trade secrets, making it difficult to enhance efficiency and productivity through digital transformation and AI in their processes.
\\ 
\null\quad  Most related research cases focus on color prediction and recipe inference systems based on structured and standardized data using NN(Neural Network)\cite{ref1, ref2, ref3}. Also, advanced algorithms like genetic, fuzzy with NN applied to infer the recipe \cite{ref4, ref5, ref6, ref7}
Although standardized and structured data is provided by global raw material suppliers, most SMEs using domestically sourced materials conduct their processes with only unstructured and non-standardized information due to productivity and cost issues. 
\\
\null\quad  In this paper, we proposed the model based on Tokenizing Encoder Neural Network (TENN) that can infer color recipes with unstructured data and tacit knowledge, which are not standardized or structured.

\section{Dataset and Model}
In this study, the RGB Color Codes Chart was used as a sample dataset for color recipe inference in view point of color engineering, such as painting, synthetic resin production, paint manufacturing, and injection molding. For data labeling, natural language text data was used as input data for each RGB color code, and the RGB color codes for each color were selected as output data. In this study, the model was trained to recognize emotional information(corpus) within the text data so that it could infer and output the RGB code combination recipes.
\\ 
\null\quad Because of the security issue and large variance of unstructured data, such as mixing ratio information of pigments and paints in SMEs, standardized sample dataset is useful in this study. The RGB color information also included subjective elements such as light, dark, pale, deep, and medium, utilizing RGB code data (composed of numbers ranging from 0 to 255 in three dimensions) for each color.
\begin{figure}[ht]
    \centering
    \includegraphics[width=15cm]{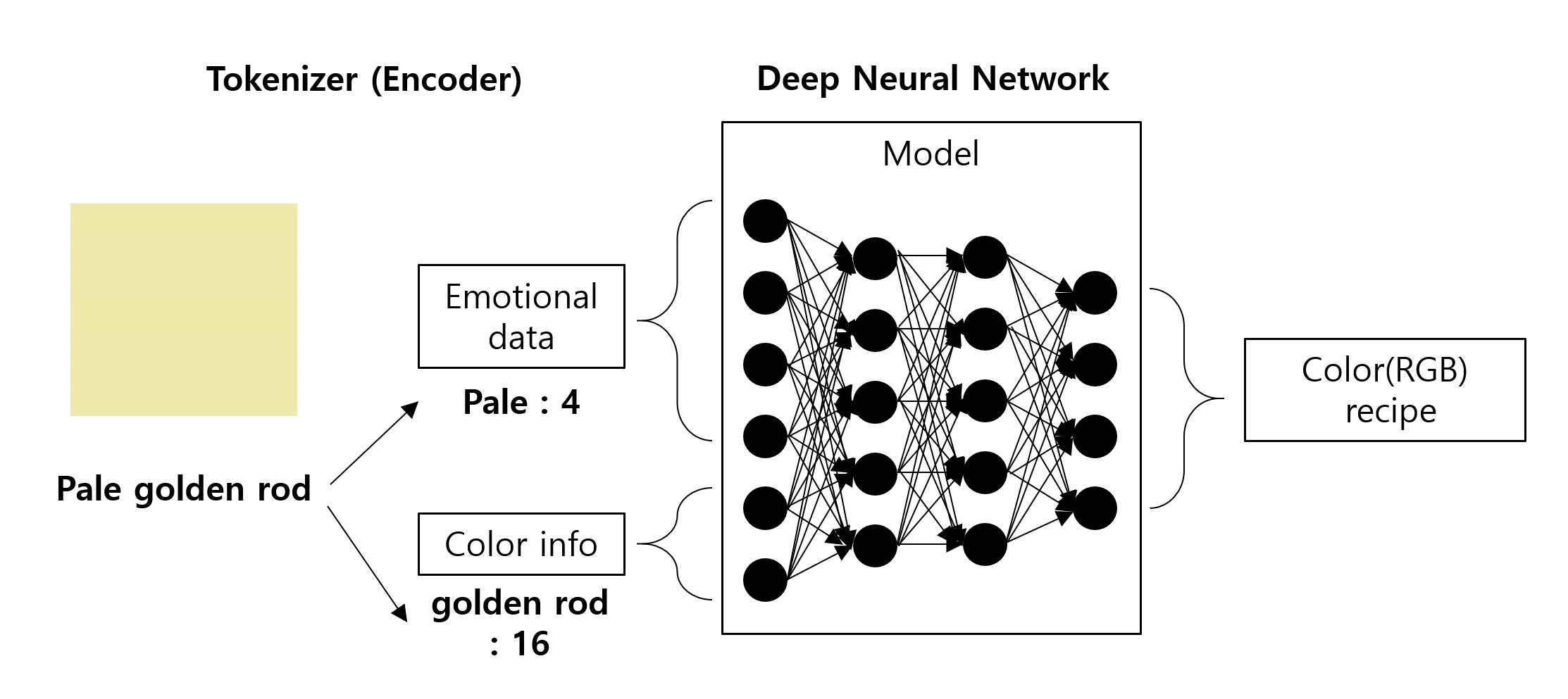}
    \caption{Model architecture}
    \label{fig:Model architecture}
\end{figure}
\\ 
\null\quad As for the model, it requires inference of untrained recipes and unstructured data based on natural language with nonlinear data patterns. Therefore, an artificial neural network with robustness and nonlinearity was applied, and a tokenizing encoder was configured as the pre-processing layer of the neural network. Existing natural language processing algorithms that have learned emotional information and unstructured data in text form through preprocessing techniques such as tokenization and encoding apply complex structures like LSTM, RNN, and those including T5, GPT, BART, and BERT. However, these structures are suitable for large-scale applications and large language models(LLM) and leading to issues with efficiency and computing resources when used in this study. There has been some research on inference systems based on emotional information using tokenization technique\cite{ref8, ref9}. However, there are few cases where this system has been applied to solve problems related to tacit knowledge scattered throughout the industry. Fig 2. shows data pre-processed results, each corpus encoded by tokenization(quantization) and matched with RGB color recipe.

\begin{figure}[ht]
    \centering
    \includegraphics[width=10cm]{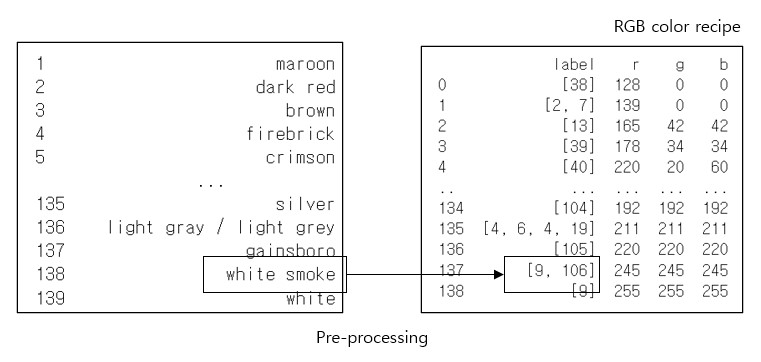}
    \caption{Data pre-processing, RGB color recipe}
    \label{fig:Model architecture}
\end{figure}

\section{Experiments}

The neural network was trained by normalizing the encoded values obtained through tokenization of the input data. To find the normalization method most suitable for the TENN model, the results of different preprocessing techniques—MinMaxScaler, MaxAbsScaler, RobustScaler, and StandardScaler— were compared on models with the same number of layers and nodes. The best method was selected based on accuracy from the learning results. The model structure consists of 12 layers with 38,731 parameters, with the first preprocessing layer performing tokenizing and encoding functions. The dataset was divided into training and testing datasets at a ratio of 0.8 to 0.2, respectively, for the experiment.
\\ 
\null\quad In this study, the Relu($  f(x)=max(0,x) $) was used as a activation function except MaxAbsScaler because of negative values in learning process. The other activation function was linear activation. In related works \cite{ref10, ref11}, the activation function can perform with it's robustness in case of non-linear systems.

\begin{table}[ht]
\centering
\caption{Results by Normalization Method}
\label{tab:my-table}
\begin{tabular}{|c|c|}
\hline
\textbf{Methods} & \textbf{Acuracy} \\ \hline
MinMaxScaler     & 0.79             \\ \hline
MaxAbsScaler     & 0.68             \\ \hline
RobustScaler     & 0.42             \\ \hline
Standardscaler   & 0.73             \\ \hline
\end{tabular}
\end{table}

\section{Results}
In this paper, we proposed and validated a color inference model using RGB codes data based on unstructured information and color information with affective corpus. We utilized a dataset of over 300 samples and confirmed the performance of the TENN. Based on the training unstructured corpus with emotional meanings, we were able to identify patterns of emotional information in the test dataset and verify whether similar patterns could be applied to other colors as a results of inference with recommend RGB color codes.
\begin{figure}[ht]
    \centering
    \includegraphics[width=15cm]{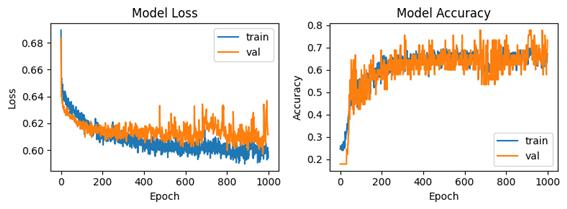}
    \caption{Model architecture}
    \label{fig:Results(loss, accuracy)}
\end{figure}
\\ 
\null\quad 
Based on the trained model, we input unstructured data in the form of color information and confirmed whether the resulting RGB recipes corresponded to similar colors. We calculated Delta E using the most recently defined standard color difference formula, CIEDE2000. The color distance averaged 73.8(Delta E value) across 30 sample data. 
\\ 
\null\quad CIELAB based Delta E is a standard measurement unit created by the Commission Internationale de l'Eclairage (CIE), which quantifies the difference between two colors and uses Lab color space transform from RGB format. A lower Delta E($  \Delta \mathbf{E} = \sqrt{(\mathbf{L_{1}-L_{2}})^{2}+(\mathbf{a_{1}-a_{2}})^{2}+(\mathbf{b_{1}-b_{2}})^{2}} $) value indicates higher accuracy, while a higher Delta E value signifies a significant discrepancy. Compared to the accuracy levels of 80\textendash 
 85 \% from previous studies\cite{ref7, ref8, ref9} on emotional information inference systems using text data from books, news, etc., our model achieved an accuracy level approximately 5\% lower.
In Fig 4, first color reference name is very light grey, and next one is cocoa brown color, each reference and model recommended color can be distinguished by naked eyes. however, view point of inference and recommend system, even it has large value of delta E, it has robustness out of range within training dataset.
\begin{figure}[ht]
    \centering
    \includegraphics[width=8cm]{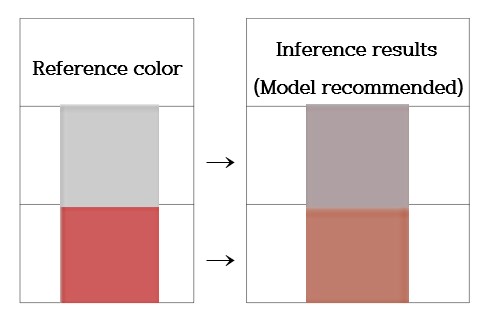}
    \caption{Model architecture}
    \label{fig:Model inference results}
\end{figure}

\section{Conclusion}
In this paper, we proposed and validated a color inference model using RGB codes data based on unstructured information and color information with affective corpus. We utilized a dataset of over 300 samples and confirmed the performance of the TENN. Also, the fact that tokenizing encoder based NN has robustness with unstructured emotional data showed positive possibility of applying advanced model including LSTM, RNN. 
\\ 
\null\quad Our future work involves utilizing pigment and formulation recipe datasets from actual painting processes used by SMEs. We plan to apply various algorithms to language models and enhance TENN by overcoming security challenges through methodologies such as federated learning and data processing with advanced algorithms. Additionally, we aim to identify potential client companies and conduct PoC as use cases.

\bibliographystyle{unsrt}  
\bibliography{ref}

\end{document}